\documentclass[letterpaper, 10 pt, conference]{ieeeconf}  % Comment this line out if you need a4paper

\IEEEoverridecommandlockouts                              % This command is only needed if 
\usepackage{graphicx}
\usepackage{subfigure}
\usepackage{tabularx}
\usepackage{dblfloatfix}
\usepackage{textcomp}
\usepackage{hyperref}
\usepackage{color}

\definecolor{NavyBlue}{rgb}{0,0,0.6}

\usepackage{graphics} % for pdf, bitmapped graphics files
\usepackage{amsmath} % assumes amsmath package installed
\usepackage{amssymb}  % assumes amsmath package installed

\title{\LARGE \bf
Shape-independent Hardness Estimation Using Deep Learning and a GelSight Tactile Sensor
}

\author{Wenzhen Yuan$^{1}$, Chenzhuo Zhu$^{2}$, Andrew Owens$^{3}$, Mandayam A. Srinivasan$^{4}$ and Edward H. Adelson$^{5}$% <-this % stops a space
	\thanks{$^{1}$Department of Mechanical Engineering, and Computer Science and Artificial Intelligence Laboratory (CSAIL), MIT, Cambridge, MA 02139, USA {\tt\small yuan\_wz@csail.mit.edu}}
	\thanks{$^{2}$ CSAIL, MIT, Cambridge, MA, USA and Department of Electronical Engineering, Tsinghua University, Beijing, 100084, China {\tt\small zhucz13@mails.tsinghua.edu.cn}}
	\thanks{$^{3}$Electrical Engineering and Computer Science Department, U.C. Berkeley and CSAIL, MIT, Cambridge, MA 02139, USA {\tt\small andrewo@mit.edu}}
	\thanks{$^{4}$Laboratory for Human and Machine Haptics (MIT TouchLab), Research Laboratory of Electronics and Department of Mechanical Engineering, MIT, Cambridge, MA 02139, USA and UCL TouchLab, Computer Science Department, UCL, London, UK {\tt\small srini@mit.edu}}
	\thanks{$^{5}$Department of Brain and Cognitive Sciences and CSAIL, MIT, Cambridge, MA 02139, USA {\tt\small adelson@csail.mit.edu}}
}

\begin{document}

\maketitle
\thispagestyle{empty}
\pagestyle{empty}

%%%%%%%%%%%%%%%%%%%%%%%%%%%%%%%%%%%%%%%%%%%%%%%%%%%%%%%%%%%%%%%%%%%%%%%%%%%%%%%%
\begin{abstract}
		Hardness is among the most important attributes of an object that humans learn about through touch.
However, approaches for robots to estimate hardness are limited, due to the lack of information provided by current tactile sensors. In this work, we address these limitations by introducing a novel method for hardness estimation, based on the GelSight tactile sensor, and the method does not require accurate control of contact conditions or the shape of objects. 
A GelSight has a soft contact interface, and provides high resolution tactile images of contact geometry, as well as contact force and slip conditions. 
In this paper, we try to use the sensor to measure hardness of objects with multiple shapes, under a loosely controlled contact condition. The contact is made manually or by a robot hand, while the force and trajectory are unknown and uneven. 
We analyze the data using a deep constitutional (and recurrent) neural network. 
Experiments show that the neural net model can estimate the hardness of objects with different shapes and hardness ranging from 8 to 87 in Shore 00 scale.	
	
\end{abstract}

%%%%%%%%%%%%%%%%%%%%%%%%%%%%%%%%%%%%%%%%%%%%%%%%%%%%%%%%%%%%%%%%%%%%%%%%%%%%%%%%
\section{Introduction}

How can we tell the difference between a stone and sponge? How can we learn if a tomato is ripe? We touch them casually, and quickly learn their hardness. 
 Humans learn a significant amount of information about the objects around them through touch~\cite{lederman1993extracting}, including hardness. 
 Hardness is defined as the resistant force of a solid matter when a compressive force is applied, or in other words, the ratio between the displacement created by an indentation and the contact force. And indeed, hardness-measuring devices such as durometers generally work by measuring the indentation produced by a known force.  Humans, however, seem to estimate hardness via a different procedure -- one that involves observing the deformation of a fingertip passively~\cite{Srini95}.  In other words, they make heavy use of cutaneous sensing, rather than kinesthetic information.

In contrast, techniques for robots to estimate object hardness are limited. 
As the majority of tactile sensors measure force, a typical trial for a robot to measure hardness is by measuring the force changes when contacting a sample with strictly controlled movement. The method is limited in that it requires strict control on both object geometry and robot manipulation.

%-----GelSight, like human finger, measures force and geometry; also aply dynamic information------
%-----Previous wok and limitation limited. question: How to understand data. We use neural network-----
\begin{figure}
	\centering{
		\includegraphics[scale=1.05]{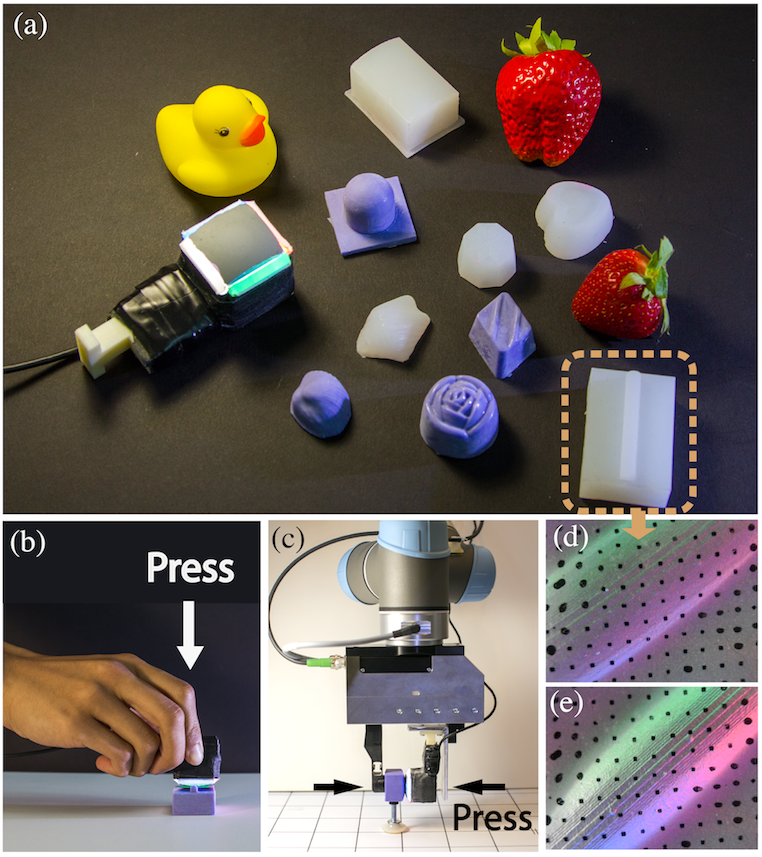}	
	}
	\caption{GelSight sensor and test samples. GelSight is an optical tactile sensor that gets high resolution tactile images of contact surface topography and approximate contact force. 
		(b) and (c) show GelSight contacting a deformable silicone sample: either a human tester presses the sensor on the object, or a robot gripper squeezes the object with the sensor. 
		The right two pictures in the bottom shows the images captured by GelSight when contacting a soft cylinder and a hard cylinder: the color indicates the surface normal on the contact surface, the black dots are markers on the sensor surface, and their displacement is related to contact force.  }
	\label{fig:IntroFig}
\end{figure}

Taking inspiration from human tactile sensing, we apply a soft tactile sensor called GelSight\cite{GelSight2009}\cite{GelSight2011}. A GelSight sensor uses a thin piece of soft elastomer to contact external objects. 
An embedded camera takes pictures of the elastomer surface, which performs shading that varies with geometry under the specially designed optical system. The image shows the contact geometry with high-resolution, as well as general contact force.
The sensor is similar to cutaneous sensing on human fingertips, in that it infers touch information from the deformation of the soft tissue. 

\begin{figure}[t]
	\centering{
		\includegraphics[scale = 1.0]{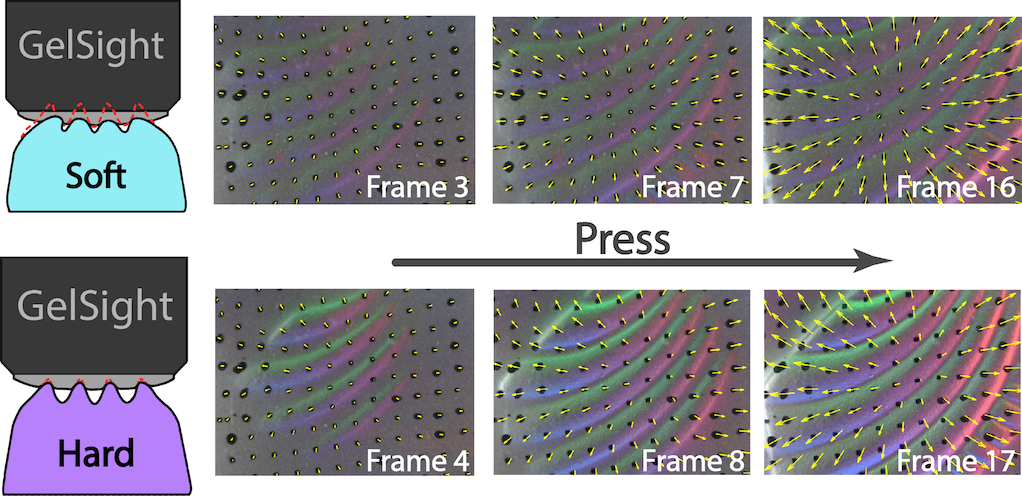}
	}
	\caption{The comparison of GelSight contacting a soft and a hard samples with the same shape. During the contact, soft object deforms more, with the ridges flattened, producing a smoother surface; hard sample deforms less, and ridges remain sharp. GelSight measures the 3D geometry of the surface, where larger height gradient causes bright color in the image, and the arrow field of black markers indicates contact force.
		We take a sequence of images during the contact procedure as our input signal.}
	\label{fig:PressExample}
\end{figure}

When the sensor is pressed against an object, they both deform according to the hardness, which results in the difference in the surface curvature and contact force. 
We showed in our previous work~\cite{GelSightIROS16} that these physical changes could be measured using simple image cues: namely, changes in intensity and motion of the black markers embedded in the gel.
Figure~\ref{fig:PressExample} shows an example. 
%In the work, we accurately estimated hardness from GelSight videos by building a model with hand-crafted features, and used them as part of a linear regression framework to estimate hardness.   

%We encountered the universal problem: how to measure the hardness of objects when they are of arbitrary, even unknown shapes? The infinite object shapes could cause very infinite response patterns of the tactile sensor, which is almost impossible to predict with a simple model. 
%To solve the problem, we apply neural network methods, which has been growingly popular due to its excellent capability in understanding large degree-of-freedom information.
While there is a strong physical motivation for these features, they make restrictive assumptions about the objects' geometry -- namely, they assume that the objects are spherical. We'd like to use this sensor with a wider variety of object types.  This, however, requires us to develop features that are applicable in more general settings.  To address this problem, we take inspiration from recent work in computer vision that learns these image features in lieu of hand-crafted features. These methods, starting with the seminal work of Krizhevsky et al.~\cite{krizhevsky2012imagenet}, have recently achieved state-of-the-art performance on many recognition tasks.  Here we apply these ideas to the domain of tactile perception by training a deep neural network to directly regress hardness directly from a raw GelSight video.   To do this, we represent frames of the GelSight video using a convolutional neural network \cite{simonyan2014very}, and we use a recurrent neural network \cite{hochreiter1997long} to model changes in the gel deformation over time.

%----Briefly tell the project outline------

In this project, we use GelSight to contact objects in a loosely controlled condition, and measure the object hardness through the sequence of tactile images from GelSight.
The sensor is either manually pressed on the samples, or mounted on a robot gripper which squeezes the samples. 
We train a neural network based on the sequences of tactile images during the presses on samples of known hardness, and predict the hardness of samples with unknown shapes. 
The test samples are made of silicone, with hardness that ranges from 8 to 87 in the Shore 00 scale, which is generally the hardness of gummy bears and pencil erasers. The samples are either of basic shapes including hemispheres and cylinders of 10 different radii, flat surface, edges and corners; or complicated shapes like everyday objects.
We also test the model on a set of natural objects like tomatoes, candies, rubber tubes. 
Experimental results show that the model can well predict the silicone samples of basic shapes, regardless of the radii of the hemispheres or cylinders. For natural objects, the model can estimate different hardness levels.

\section{Related Work}

\subsection{Optical Tactile Sensors and GelSight}

Over the past decades, researchers have proposed many techniques for tactile sensing~\cite{TactileReview2010} in robotic applications.  Among these, piezoresistive, peizoelectric, and capacitive sensors that measure the local force distribution have emerged as the dominant approaches.  While tactile sensors based on these techniques have been used successfully in many applications, they often result in sensors that are difficult to fabricate, and which have limited spatial resolution or area.  Optical-based tactile sensors~\cite{Ferrier2000,GelForce2005,Bristol2009}, on the other hand, overcome many of these limitations.  Most of those optical tactile sensors use a piece of deformable material as contact medium, and add some visually trackable patterns or markers to the medium. The sensors then infer the deformation of the contact medium by observing the deformation of the patterns, thereby measuring the contact force. 
%Optical based tactile sensors usually can well locate the load over a relative large area, but it is usually challenging to accurately measure the force distribution on an arbitrary surface. 

In this work, we consider the GelSight touch sensor -- an optical tactile sensor that is designed to measure high-resolution shape of the contact interface. It consists of a clear elastomeric slab with a reflective membrane on the surface, with an embedded camera and illumination system~\cite{GelSight2009, GelSight2011}. When the membrane is in contact with an object, one can obtain a highly accurate height map (accurate within microns in \cite{GelSight2011}) of its deformation via photometric stereo. 
This sensor has been used to measure physical properties in many applications.  Jia, et al.~\cite{GelSightLump}, for instance, showed that it was more accurate than human subjects in detecting hard lumps in soft tissues, and Li~\cite{GelSightUSB} proposed a fingertip GelSight device, which is small enough to be mounted on a robot fingertip. 
Yuan et al.~\cite{GelSightShear} added markers to the GelSight surface and provided a method for tracking their movement, thus enhancing its ability to infer the approximate contact force on the sensor.  

%% With a GelSight sensor, a robot would not only be able to quickly determine the shape and texture of an object, but also infer more physical properties that remain to be explored.  

%In this paper, the proposed method begins with a qualitative observation: the tactile image sequences from GelSight look quite different for hard objects versus soft ones. When a hard object is pressed, it retains its shape as force increases. In contrast, a soft object's shape flattens out as force increases, and the depression is distributed more uniformly over the contact area. In addition, the boundary of the contact region is more pronounced for a hard object than for a soft one. This suggests that it is possible for a GelSight sensor to estimate hardness even when the force is unknown or poorly controlled, as may occur in the real-world situations. 

\subsection{Hardness Measurement}

Robotics researchers have long been interested in estimating object properties through active touch. Drimus et al.~\cite{kragic2011} and Chu et al.~\cite{kuchenbecker2013}, for example, recorded a sequence of contact force signals and then inferred object properties from change in force. These methods often implicitly include hardness in the signal. However, research on  directly estimating hardness from a tactile sensor has been more limited.  

%% touch sensors---------Consider the precision
For a robot, the most straightforward way to measure object hardness is by applying controlled force and measuring deformation, or vice versa~\cite{HapticCues2009}. 
Su et al.~\cite{BioTacCompliance} measured the hardness of flat rubber samples using a BioTac touch sensor, which measures multiple tactile signals including the pressure on fixed points. In the work, the sensor was installed on a robot fingertip, and it is pressed onto flat silicone samples in a strictly controlled motion. Changes in the force re then used to discriminate between 6 samples with different hardness values ranging from 30 Shore00 rubber to rigid aluminum. For this method to be applicable, however, the sensor movement and object geometry must be strictly controlled. 
In contrast, our input is more general; we directly estimate hardness (rather than distinguishing between samples); and we use more diverse geometries and hardness.

Another specific way is to design a touch sensor that has special mechanisms for explicitly measuring both the local force and the depth of the press. For example, Shimizu et al.~\cite{shimizu2002sensor} designed a piezo-resistant cell with a gas-filled chamber, which was used to measure the indentation of the mesa on its top surface. The cell thus measures the material's hardness from the force measured by pressure change in the chamber and the indentation depth measured by piezo-resistance. The sensor makes measurement easier, under the condition that the surface geometry is certain.  These limitations constrain the use of the sensor for more general touch tasks. There are also sensors designed specifically for measuring hardness for medical uses, such as \cite{omata1992new}\cite{SkinHard98}, or with ultrasonic signals instead of force signals~\cite{takei2004}. 
However, these sensors are not designed for measuring hardness of general objects, and are not well suited to other tactile tasks. 

%----what we have done

In our previous work~\cite{GelSightIROS16}, we showed that GelSight can accurately estimate the hardness of a set of hemispherical silicone samples under loosely controlled contact conditions. As in this work, we asked a human tester and an open-loop robot to press on samples, and developed a model to predict the hardness according to both the changing brightness in the GelSight image and the displacement field of the surface markers. While the model successfully predicted the hardness, it was limited to hemispherical samples, and it required that the radii  be provided as input.

\subsection{Learning for tactile sensing}

Recently, researchers have proposed several learning-based methods
that estimate tactile properties from touch sensors.  Gao et
al.~\cite{gao2015deep}, for example, used a neural network to infer
haptic adjectives (i.e., qualitative properties such as "bumpy" or
"squishy") from a Biotac sensor.  While their technical approach also
uses a recurrent neural network, specifically a 1D convolutional
network and recurrent network, the output of the sensor-- an
assortment of 32 time-varying physical measurements, such as fluid
pressure and temperature -- is substantially more limited. In
contrast, our method starts with an input that is significantly lower
level -- namely, images of a deforming gel.  From this rich signal, we
infer metric hardness measurements, rather than qualitative properties.

%Our previous work~\cite{GelSightIROS16} performed this task by
%defining physically motivated image features and regressing hardness
%from these features.  However, since the input is simply a sequence of
%images, we can analyze it using standard computer vision models that
%learn end-to-end, directly mapping from pixels to tactile properties.
%Our model resembles methods such as \cite{donahue2014long} that have
%been successfully applied to action recognition: We represent the
%GelSight images using a deep convolutional neural network (CNN)
%\cite{krizhevsky2012imagenet,lecun1989backpropagation}. Specifically,
%we use the 16-layer VGG architecture \cite{simonyan2014very} which has
%been shown recently to achieve high performance on large-scale object
%recognition tasks.  Then, to represent the temporal changes in the
%signal, we use a recurrent neural network with long short-term memory
%units (LSTM) \cite{hochreiter1997long}.  Using this network, we
%directly regress a measurement of hardness.

\section{Estimating Hardness from GelSight Videos}

\begin{figure}
	\centering{
	\includegraphics[]{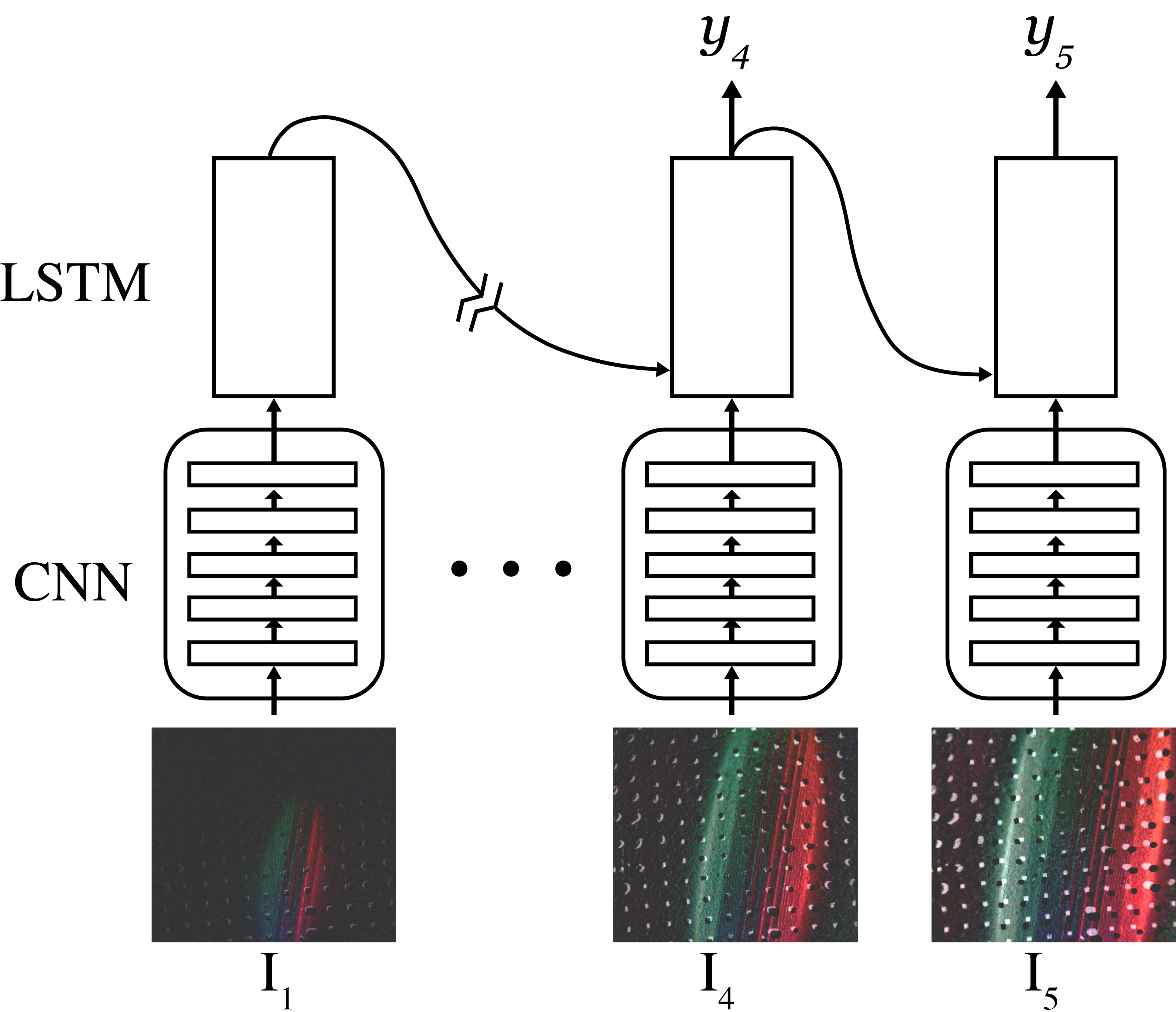}		
		}
\caption{
	We use a recurrent neural network to map a video recorded by the GelSight sensor to a hardness value.  The network resembles \cite{donahue2014long}: images are represented using CNN features {\em fc7} from a VGG16 net, and feed into an LSTM net.
The net input $I_t$ is a sequence of GelSight image after subtracting the initial frame. In particular, we choose 5 frames evenly from a pressing sequence, and each frame is of one pressing stage. We only use the RNN output of the last 3 frames, i.e. $y_3, y_4, y_5$ to estimate the sample hardness. }
\label{fig:net}
\end{figure}

Figure \ref{fig:PressExample} shows the examples of videos recorded by a GelSight sensor.  Here, the differences between hard and soft objects are readily apparent: a hard object deforms little during the contact, while a softer object may deform largely and makes a flatter surface under relatively smaller force. 
These subtle deformations in shape and force, in turn, are visible in the videos recorded by GelSight.  
As in \cite{GelSightIROS16}, the GelSight video shows the surface normal of the contact surface through the inner surface's reflection change under the colored and directional light sources, which results in different color intensity.  
Furthermore, the black markers embedded on the surface help convey the contact force through their motion.

\subsection{Neural network design}

We ask in this work whether we can use these cues to predict an arbitrary object's hardness. To do this, we use a neural network that maps an image sequence to a scalar hardness value (measured on the Shore 00 scale).  The design of the network is similar to recurrent networks considered in action recognition~\cite{donahue2014long}, and is shown in Figure~\ref{fig:net}. We represent each GelSight image $I_i$ with convolutional network features $\phi(I_i)$. For these, we use the penultimate layer ({\em fc7}) of the VGG architecture\cite{simonyan2014very}.  We then use a recurrent neural network with long short-term memory units (LSTM) \cite{hochreiter1997long} to model temporal information. 

At each timestep, we regress its output hardness value via an affine transformation of the current LSTM's hidden state $h_t$:
\begin{eqnarray}
y_t &=& Wh_t + b \nonumber\\
h_t &=& L(h_{t-1}, \phi(I_t)),
\end{eqnarray}
where $W$ and $b$ define an affine transformation of the hidden state $h_t$,
and $L$ updates $h_t$ based on the previous state $h_{t-1}$ using the current image $I_t$ (here we omit
the LSTM's hidden cell state for simplicity).

The prediction $y_t$ is the hardness estimate for the current timestep. We estimate a hardness value for the object as a whole by averaging the predictions from the final 3 frames. We perform the regression on a per-frame basis to add
robustness to videos in which the pressing motion differs significantly from those of the training set. During training, we minimize a loss that penalizes the difference between the predicted and ground-truth hardness values, using a Huber
loss. 

\subsection{Choosing input sequences}
We'd like to make our method invariant to the speed of the pressing motion and to the maximum contact force. 
For example, different human testers or robots may manipulate objects with different loading speed or maximum force.
Therefore, we constrain the video sequence so that it
begins and ends at times that are consistent across manipulation
conditions. Specifically, we choose a 5-frame sequence of images in the loading period of the press,
such that the sequence starts after the object being pressed. We
determine this starting point by finding the frame in which the mean
intensity of the GelSight image (a proxy for the force of the press)
exceeds a threshold.  For the end point of the sequence, we choose 
the last frame whose intensity change is the peak of the sequence.
The other 3 frames are evenly chosen in the middle according to the intensity change.

We subtract the first frame in the sequence from the chosen frames to
account for preexisting deformations in the elastomer. Unlike in
\cite{GelSightIROS16}, we do not explicitly model the motion of the
markers that are embedded in the gel -- relying instead on the network
to learn about their motion via the raw images (in which the motion of
the marker is implicitly visible due to subtracting the initial
frame).

\subsection{Training}

\begin{figure*}[h]
	\centering{
		\includegraphics[scale = 1]{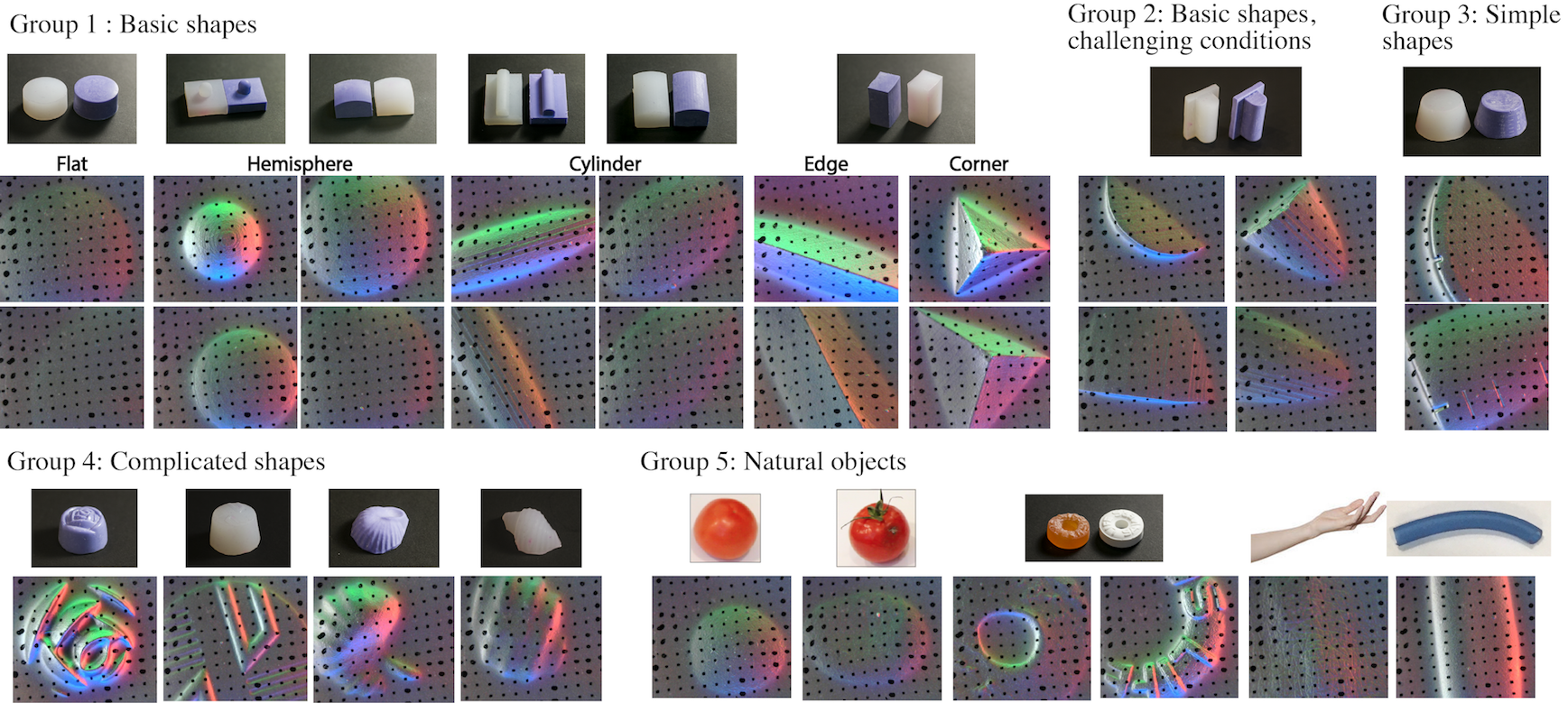}
	}
	\caption{Examples of GelSight dataset. Group 1 are of the basic shapes, where the second row is of the tactile images of softer samples with same shape; Group 2 are of contacting the same set of samples, but in a undesired contact condition; Group 3 are the same of some natural object shapes but with simple geometry; Group 4 are complicated shapes made by chocolate molds; Group 5 are natural objects, and in this figure the left two are from tomatoes, right two are from candies. }
	\label{fig:Dataset}
\end{figure*}

The training dataset contains about 7000 videos obtained by having human testers press GelSight on different silicone samples, each video is an independent pressing sequence. 
The training dataset contains mainly the basic object shapes(Group 1 in Figure~\ref{fig:Dataset}), but also with a large portion of complicated shapes or bad contact conditions(Group 2 and 4 in Figure~\ref{fig:Dataset}). Those irregular data greatly help to prevent overfit of the model.
A single video is used for multiple times during the training, with different end point for sequence extract, so that the contact situations with different maximum forces are included.
In other words, when choosing a sequence ends at the middle of the loading process, the sequence equals to the case of pressing on the same object in the same way, but with smaller maximum force.  
We train the model using stochastic gradient descent, initializing the CNN weights with ImageNet
\cite{krizhevsky2012imagenet} pretraining, jointly training the CNN
and LSTM.  We train the algorithm for 10,000 iterations, a learning rate of 0.001, and a step size of 1000.

\section{Experimental Setup}

%--- GelSight
\begin{figure}[t]
	\centering{
		\includegraphics[height=2 in]{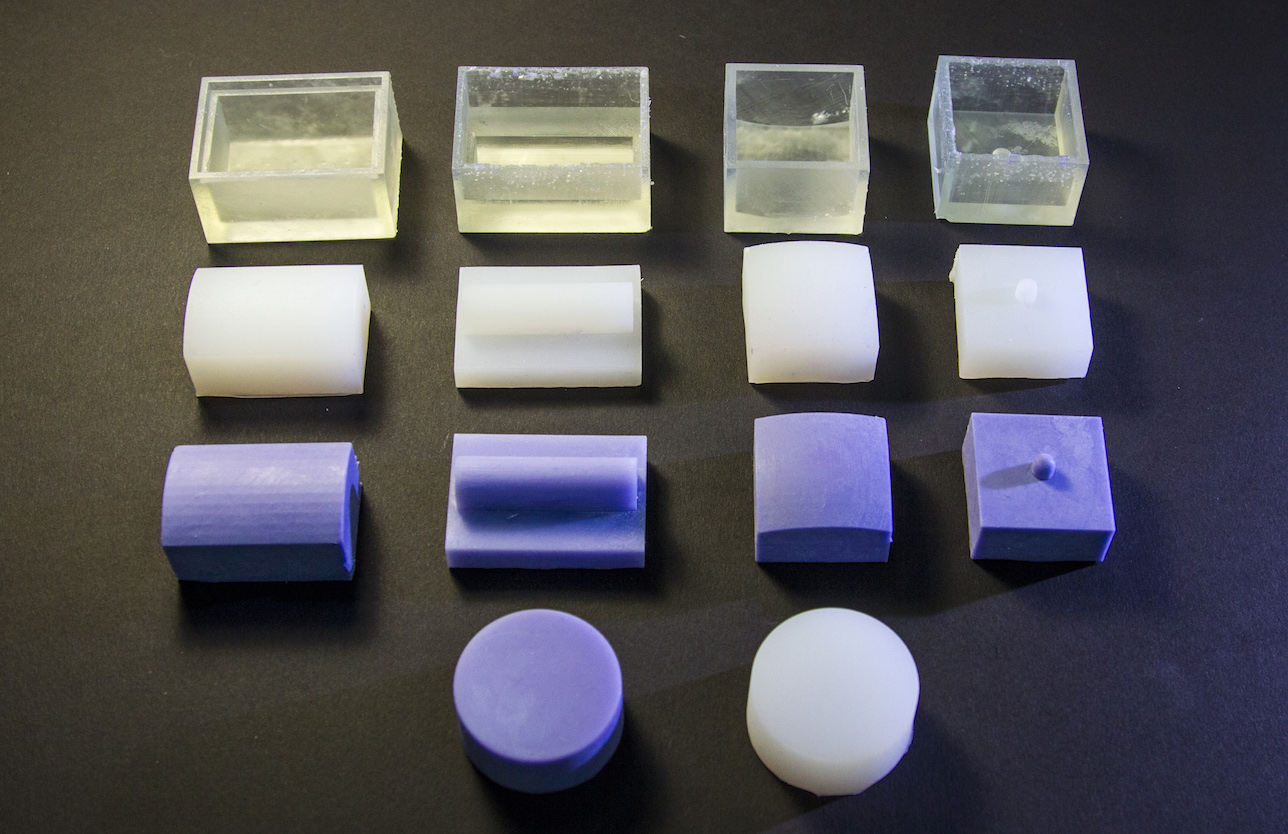}
	}
	\caption{Some 3D printed molds and silicone samples in hemispherical and cylindrical shapes. Samples are of different radii and hardness, and hardness is controlled by changing silicone ingredient mixing ratio. 
	The material hardness is tested by a durometer on standardized silicone samples, as the bottom two samples.
	}
	\label{fig:SampleMold}
\end{figure}

The touch sensor we used is a Fingertip GelSight device ~\cite{GelSightUSB}, with black markers on the surface to track the displacement field. The elastomer on GelSight is of the dome shape, with the area of $25\mathrm{mm}\times 25\mathrm{mm}$, and the maximum thickness in the center is around $2.4\mathrm{mm}$. The average interval between the black markers is about 1.3 mm. 
The hardness of the elastomer is 17 in Shore 00 scale. The camera in the sensor takes original images of size $960\times720$ pixels over an area of $18.4\mathrm{mm}\times 13.8\mathrm{mm}$ at a speed of 30Hz. 

%--- How to make the samples; List of Samples; Picture of sample and molds; Parameters of the samples; We control the overall height of the samples

The silicone samples that we used in our experiments have the varied shapes of 1) hemispheres of different radii, 2) cylinders of different radii, and 3) some arbitrary shapes. The silicone materials were Ecoflex\textsuperscript{\textregistered} 00-10, Ecoflex\textsuperscript{\textregistered} 00-50 and Smooth-Sil\textsuperscript{\textregistered} 945 from Smooth-On Inc. The three raw materials are mixed by different ratio to make the samples into 16 different hardness levels between 8 in Shore 00 scale to 45 in Shore A scale (equals to 87 in the Shore 00 scale). The silicone materials are in liquid phase, and they will solidify in several hours after Part A and Part B are mixed and poured into the mold.
% Shape of the samples
The hemispherical and cylindrical samples have 9 different radii ranging from 2.5mm to 50mm, but they are of similar height (around 25mm). This design is to ensure the sample thickness would have limited influence of on the shape change during the press. We fabricate the mold with a Form 2 3D printer from Formlabs.
For the arbitrary shapes, one group is of simple and common shapes, that is casted from some daily vessels, like square shaped ice box, truncated cone shaped measuring cups, small beakers; another group are of complicated and special shaped, casted from assorted chocolate molds. 
They include the shapes of different emboss textures or complicated curvatures, like the shape of shells. 
%we choose some sorted chocolate molds for sample casting. The molds provide a wide set of shapes, including curved smooth surface, flat emboss patterns, thin cylindrical patterns, edges of different curvatures, and curved surface patterns. 
In total, we made 95 hemispherical samples, 81 cylindrical samples, 15 flat samples and 160 samples of arbitrary shapes for experiments. We also collect a set of 20 rigid objects in the database. 

%----- How to get the ground truth; Durometer

The sample hardness is measured by a PTC\textsuperscript{\textregistered} 203 Type OO durometer. 
%The durometer measurements require the test samples to be of flat shape, and have a thickness of over 6mm. 
We made a set of flat thick silicone samples(the bottom two samples in Figure~\ref{fig:SampleMold}) of each mixing portion as the test sample from durometer test. To reduce the measurement error, we took 5 tests and use the mean value for each test sample . 
The real samples are considered as of the same hardness of the samples being tested. 

%---Computer?

The hardness of the elastomer on the GelSight sensor may influence the sensitive range of the sensor. The sensor can better estimate the objects' hardness when it is closer to the sensor's hardness. In this project, we choose a soft elastomer of 17 in Shore 00 scale, in order to get a better discrimination of the very soft objects, but the sensor cannot well differentiate the hardness when it is harder than 70 in Shore 00 scale. 
\\

\section{Experimental Procedure}
\label{chpt:ExpProcedure}

When we press the GelSight sensor into a sample object, we push it in the normal direction. We conducted two sets of experiments: human testers pressing the sensor, and a robot gripper squeezing on the samples.
In the human testing scenario, the test object is placed on a flat hard surface, and a tester holds the GelSight sensor and presses on the object vertically, as shown in bottom left figure in Figure~\ref{fig:IntroFig}.
For the robot test, we used a Weiss WSG 50 gripper which has GelSight as one finger. 
The robot gripper closes in a slow and constant speed until the gripping force reaches the threshold, making the GelSight sensor squeezing on the object. The speed was randomly chosen between 5 to 7 mm/s, and the gripping force threshold is a random between 5 to 9N. 
The procedure is shown in the bottom middle figure in Figure~\ref{fig:IntroFig}.
In both cases, the sensor is pressed into the object, the contact force grows, and the deformation of both the GelSight elastomer and the object increases. We record the GelSight video during the press. In average, there are 20 to 30 frames in the pressing period.

Note that in both experimental sets, the contact condition is highly varied. In the human tester experiment, the trajectory and pressing velocity are both unknown and uncontrolled, and small amount of shear force and torque exists in the contact. 
Our goal was to model a ``natural'' tactile interaction -- similar to what one would do in daily environments, or when a robot with a touch sensor contacting an arbitrary object in a complicated environment.

\begin{figure*}[h]
	\centering{
		\subfigure[Seen shapes with unseen hardness]{\includegraphics[height=1.6in]{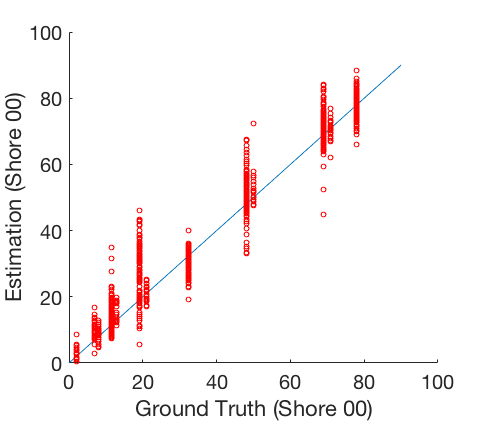}}
		\subfigure[Unseen shapes]{\includegraphics[height=1.6in]{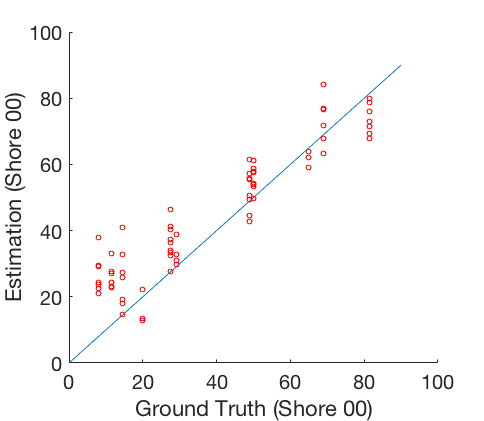}}
		\subfigure[Data collected by robot]{\includegraphics[height=1.6in]{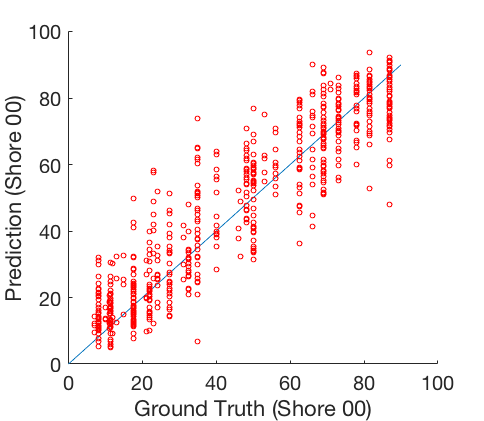}}
	}
	\caption{Prediction results on samples of basic shapes, including hemispheres, cylinders, flat surface, edges, corners.}
	\label{fig:result_basic}
\end{figure*}

The collected GelSight data are of the following geometry types, as shown in Figure~\ref{fig:Dataset}:
\begin{enumerate}
	\item \emph{Basic shapes}. Samples of the simplest geometries, including flat surface, spherical surface, cylindrical surface, straight sharp edges, sharp corners. Spherical and cylindrical shapes are of 10 levels of radii from 2.5mm to 50mm.
	\item \emph{Bad basic shapes, challenging contact conditions}. The contact objects are the same as in the previous group, but the contact condition is undesired. For example, the sensor contacts the silicone samples in tilted angles, or the sample is included in the contact area.
	\item \emph{Simple shapes}. The samples are made of silicone with known hardness and basic shape, such as shapes of frustum measuring cup.
	\item \emph{Complicated shapes}. The samples are made of silicone but with complicated textures or shapes. They are made from the chocolate molds. 
	\item \emph{Soft objects in everyday life}. These are the soft objects in the everyday life, mostly with relatively simple shapes. Human can roughly feel whether they are `soft', or `very soft', or `hard'.
\end{enumerate}

We selected some samples in Group 1, Group 2, and Group 4 as the training set, and tested the model's prediction on Group 1, 3, 4 and 5. The data in the training set is the data collected by human testers; in the test set, some data is collected by human testers, while  data is collected by a robot. The dataset is published at \url{http://people.csail.mit.edu/yuan_wz/hardnessdataset/}.

\section{Experimental Results}
\label{chpt:expres1}

\subsection{Basic Shapes}

In the first experiment, we wanted to test whether a model could generalize to new hardness values.  Therefore, the objects in the test set had the same shape as those in the training set (Group 1 in Figure~\ref{fig:Dataset} and Section ~\ref{chpt:ExpProcedure}]), but with different hardness ratings.
 In the second experiment, the samples are of cylindrical or spherical shapes, but are not included in the training set; in the third group, the samples are of the Group 1 basic shapes that have been seen, but the experiment is conducted by the robot, which makes different contact motion than human testers. 
The estimation result of the three groups is shown in Figure~\ref{fig:result_basic} and Table~\ref{tb:Res1}.

\begin{table}
	\begin{center}
		\caption{Network prediction on basic shapes}
		\label{tb:Res1}
		%\begin{tabular*} {0.45\textwidth} {@{\extracolsep{\fill}}| c |c|c|c| }
		\begin{tabular}{|c|c| c |c| }
			\hline
			 & number of videos &$R^2$ & RMSE \\
			\hline
			\hline
			Trained shape, novel hardness &1398& 0.9564  & 5.18 \\
			\hline
			Novel shapes &73& 0.7868 & 11.05\\
			\hline
			Trained samples, robot gripper&683& 0.8524 & 10.28\\
			\hline
		\end{tabular}
	\end{center}
\end{table}

\begin{figure*}
	\centering{
		%	\subfigure[]{\includegraphics[height=1.65in]{Figure/result/ltomato_bar}}
		%	\subfigure[]{\includegraphics[height=1.65in]{Figure/result/tomato_bar}}
		\includegraphics[]{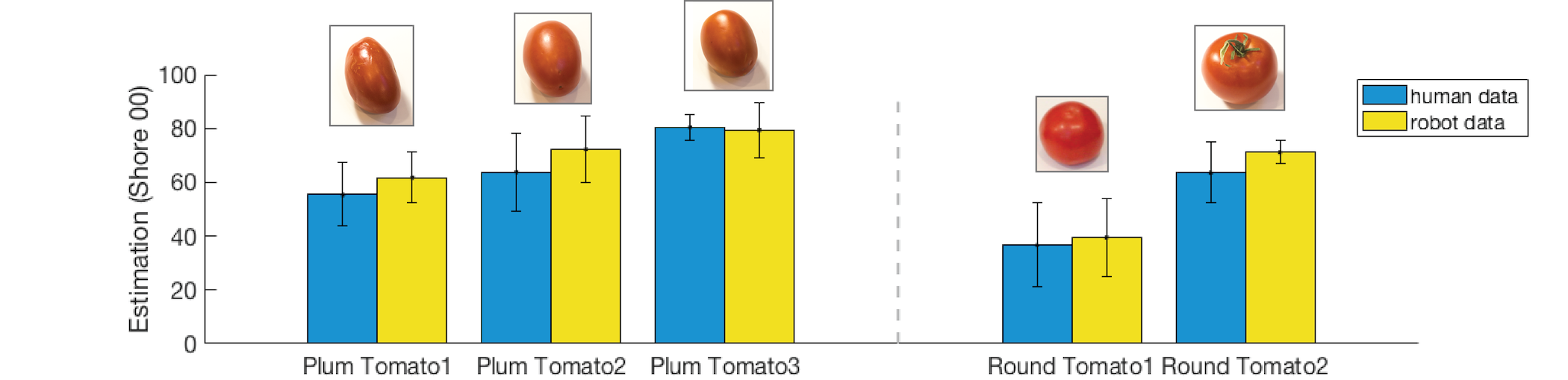}
	}
	\caption{GelSight estimation of tomato hardness, experiments executed by either human or robot. For each tomato shape, the display ranking is according to human estimation: larger index indicates harder and rawer tomatoes. In average, each tomato is pressed for 18 times on different parts, and the hardness on different parts is slightly different.}
	\label{fig:res_fruit}
\end{figure*}

\begin{figure*}
	\centering{
		%		\subfigure[]{\includegraphics[height=1.6in]{Figure/result/candy_bar}}
		%		\subfigure[]{\includegraphics[height=1.6in]{Figure/result/tube_bar}}
		%		\subfigure[]{\includegraphics[height=1.7in]{Figure/result/other_bar}}
		\includegraphics[scale = 1.1]{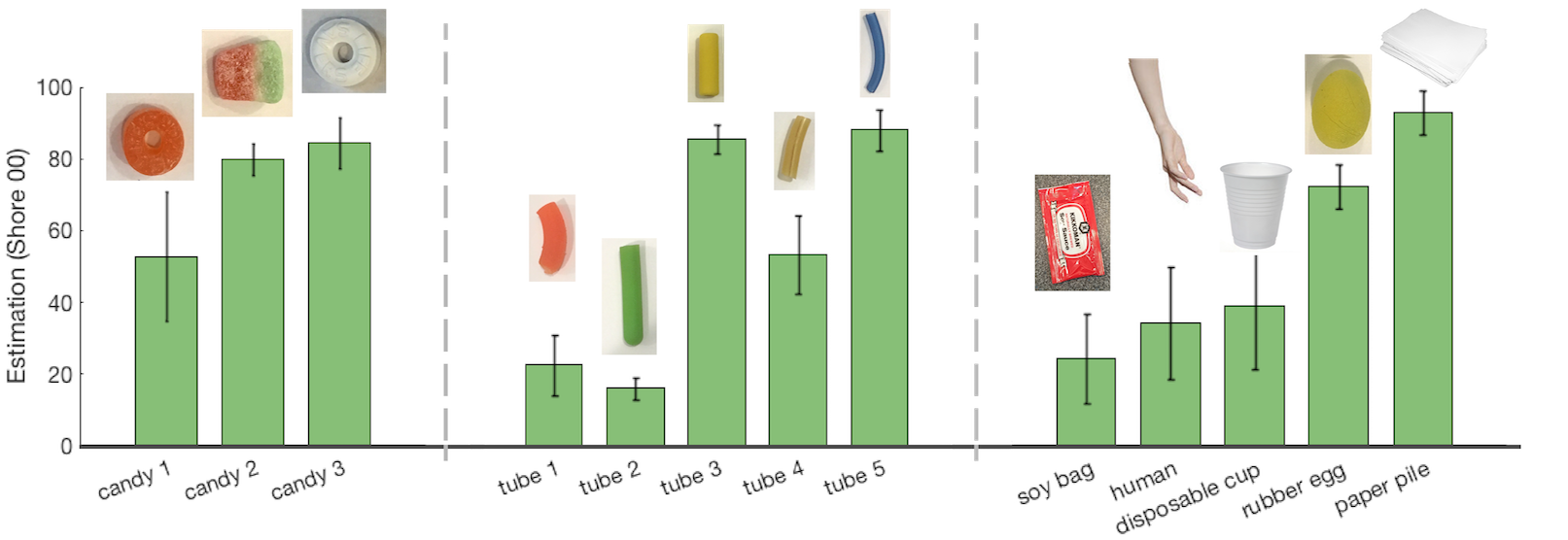}
	}
	\caption{Measurement of some natural objects. In average, each object is pressed on for 5 times by a human tester. In each group, the object order on x axis corresponds to human ranking of hardness. In the first group, candy 1 is a soft gummy candy, candy 2 is a hard gummy candy with rigid sugar particles on the surface, candy 3 is a rigid candy. Hardness of candy 2 is over estimated because it has dense particles on the surface. Similarly, in the 2nd group, hardness of tube 1 and tube 3 is overestimated because they have textures, like a ridged surface. }
	\label{fig:res_object}
\end{figure*}

\subsection{Arbitrary Shapes}
\label{chpt:expres2}

In this challenging experiment, we test our model on the arbitrary silicone shapes that the algorithm has not seen during training, as mentioned as Group 3 and Group 4 in Section~\ref{chpt:ExpProcedure} and Figure~\ref{fig:Dataset}.  
For Group 3 samples, the $R^2$ of measurement is 0.57, and RMSE is 19.3; for Group 4 samples, the $R^2$ decreases to 0.39, and the RMSE is 18.2.
In most large-error measurement cases, the model tends to estimate the hardness much higher than the ground truth. In these cases, the objects mostly have some sharp surface curvatures but not the same as as those in the training set. 
Moreover, there may be multiple ridges present in a sample.
The network tends to consider the object of higher hardness when there are sharp curvatures on the contact surface.
On the other hand, for the object shape that is included in the training set, the neural net can well estimate the target hardness.

% why this happens
when the training set contains basic shapes, the model can generalize to more complex shapes -- albeit with decreased performance.
For the method of deep neural network, a tough learning through all the possible data is required. To make the network more generalized in estimating the hardness for objects with more arbitrary shapes, a much larger training set that contains more changes in the sample shapes is required.

\subsection{Estimation of Natural Objects}

We use GelSight to contact natural objects and use the model to estimate the hardness of them. As it is difficult to measure their ground-truth hardness with standard method, we ask human subjects to rank the hardness of similar objects, and compare the readings from GelSight measurement. 
We compared the estimated hardness of several plum tomatoes and round tomatoes with different ripeness, as shown in Figure~\ref{fig:res_fruit}. Although the tomatoes are of close hardness, the GelSight prediction is similar to human estimation. 
The GelSight measurement of different candies, elastomer tubes, and some random daily objects are shown in Figure~\ref{fig:res_object}.

From the figures we can see that, for natural objects of simple geometry and smooth surfaces, our network can well estimate their hardness level. The estimation can be used to differentiate ripeness levels of some fruits, like tomatoes. Rigid objects will be estimated with a number larger than 80. 
However, similar to the experiments in Section~\ref{chpt:expres2}, the geometry of the natural objects may also have an influence on the measurement, in that the objects with textures, like a ridges surface, tend to be estimated harder. The reason is mainly due to the incompleteness of the training set for the network.

\section{Conclusion}

In this paper, we proposed a deep neural network model to estimation hardness of objects with multiple shapes using a GelSight tactile sensor. We mainly considered the commonly seen shapes of natural objects, including spheres, cylinders, flat surface, edges and corners. The GelSight sensor records the objects' deformation during the contact process in high-resolution image sequences, which contain information about both the shape change and contact force. 
The biggest challenge lies in that the different geometry of the object may have very complicated influence on the sensor output, and is nearly impossible to model with linear relationships. 
To solve this problem, we apply a convolutional neural network and recurrent neural network to extract the hardness information from the GelSight video sequence. Experimental results show that the network can well predict hardness of silicone samples with similar shape in the dataset, regardless of the loading conditions; for the objects with complicated shapes or ridged surface, the model does not estimate the hardness well, which caused by the incompleteness of the training data set. 
For many natural objects with smooth surface and simple geometries, the model can roughly measure their hardness level, thus can be used for recognize objects with special hardness, or choose fruits with preferable ripeness level.

\addtolength{\textheight}{-14.5cm}   % This command serves to balance the column lengths

%%%%%%%%%%%%%%%%%%%%%%%%%%%%%%%%%%%%%%%%%%%%%%%%%%%%%%%%%%%%%%%%%%%%%%%%%%%%%%%%

%%%%%%%%%%%%%%%%%%%%%%%%%%%%%%%%%%%%%%%%%%%%%%%%%%%%%%%%%%%%%%%%%%%%%%%%%%%%%%%%

%%%%%%%%%%%%%%%%%%%%%%%%%%%%%%%%%%%%%%%%%%%%%%%%%%%%%%%%%%%%%%%%%%%%%%%%%%%%%%%%

\section*{ACKNOWLEDGMENT}
The work is supported by a grant from Toyota Research Institute, a grant from Shell to EHA and ERC-2009-AdG 247041 to MAS. Chenzhuo Zhu gratefully thank the Spark Program and the Top Open Program at Tsinghua University.

%%%%%%%%%%%%%%%%%%%%%%%%%%%%%%%%%%%%%%%%%%%%%%%%%%%%%%%%%%%%%%%%%%%%%%%%%%%%%%%%

\bibliographystyle{IEEEtran}
\bibliography{Ref_Hardness}
%{\small
%\bibliographystyle{plain}
%\bibliography{Ref_Hardness}
%}

\end{document}